# Fuzzy Geometric Relations to Represent Hierarchical Spatial Information

**Stéphane Lapointe**
Command and Control Division
Defence Research Establishment Valcartier
2459 Blvd. Pie XI north
P.O. Box 8800, Courcelette (Québec)
Canada, G0A 1R0
lapointe@cc.drev.dnd.ca

**René Proulx**
APG Inc.
70 Dalhousie street, suite 320
Québec (Québec)
Canada, G1K 4B2
rproulx@cc.drev.dnd.ca

## Abstract

A model to represent spatial information is presented in this paper. It is based on fuzzy constraints represented as fuzzy geometric relations that can be hierarchically structured. The concept of spatial template is introduced to capture the idea of interrelated objects in two-dimensional space. The representation model is used to specify imprecise or vague information consisting in relative locations and orientations of template objects. It is shown in this paper how a template represented by this model can be matched against a crisp situation to recognize a particular instance of this template. Furthermore, the proximity measure (fuzzy measure) between the instance and the template is worked out — this measure can be interpreted as a degree of *similarity*. In this context, template recognition can be viewed as a case of fuzzy pattern recognition. The results of this work have been implemented and applied to a complex military problem from which this work originated.

## 1 INTRODUCTION

Representing spatial information by common patterns is a problem often encountered in various domains of computer science such as computer vision, pattern recognition, and spatial information systems. Patterns represent generic arrangements of various objects in space. Once these patterns are built, they can be used to recognize particular arrangements from observed data. In various applications, spatial information is hierarchically structured, that is, patterns are described in terms of other sub-patterns. Furthermore, to allow variability in patterns, it is of great interest to associate degrees of uncertainty with patterns.

These issues, related to spatial information, are the subject of this paper. The concept of spatial template is introduced to capture the idea of arrangements of objects in space (patterns). Based on the observation that patterns are often highly geometric, an expressive representation model using fuzzy geometric relations is developed. The model allows for the representation of hierarchically structured spatial information. Fuzzy set theory (Zadeh, 1965) is used to associate degrees of uncertainty with particular cases of arrangement (called template instances). The fuzzy variables are relative location and relative orientation in space. To combine the uncertain information while being consistent with operations on fuzzy sets, the minimum over various degrees of membership is taken. An algorithm that recognizes template instances from observed data and associates a degree of similarity (fuzzy measure) has been developed and implemented. The results of this work are applied in the military domain for the representation of doctrinal deployment templates of units in two-dimensional space and the recognition of instances of templates from observed situations.

Spatial template representation and recognition can be viewed as particular cases of fuzzy pattern description and recognition (Kandel, 1982). To the best of our knowledge, few previous works tackled problems similar to the one presented here. Woods (1993a, 1993b) studied the same problem using a different approach. His concern was to efficiently search the space of potential solutions using strategies for constraint satisfaction problems (CSP). He did not focus on the representation model (although he studied the problem of constraint hierarchies) and did not handle uncertainty. Another author (du Verdier, 1993) addressed a similar problem also using CSP. Similar to our approach, he used geometric constraints which can be viewed as geometric relations, however the focus of his work was not the representation model. Dumouchel (1990) built a system that provides a degree of similarity for an observed arrangement of objects with an ideal template using a probabilistic function. His method is not much automated since the user has to provide the location, orientation, depth, and width of the template before the probabilistic function can be computed. A parallel implementation for model matching that takes into account various distortions (rotation, global contraction and global expansion) has been proposed (Rigoutsos and Hummel, 1992) but object orientation, uncertainty handling and local distortions were not considered. With respect to previous researches, the main contribution of this work lies in the flexible representation model which takes into account a form of uncertainty (based on fuzzy sets theory) on locations and orientations of objects, allows various template distortions (local and global distortions), and is well suited for representing hierarchically structured spatial information.

Section 2 defines the basic concepts used in this paper. Section 3 presents the representation model which is based on fuzzy geometric relations. The recognition algorithm and implementation are discussed in section 4. In section 5, a military application and a complex example are detailed. Although we have only applied the results of this work to a military problem, we believe that other domains could benefit from these results as representation of spatial information is ubiquitous in information systems. Finally, areas for further research are proposed in section 6.

## 2   BASIC CONCEPTS

### 2.1   SPATIAL TEMPLATE

This section explains concepts related to spatial template. Informally, a spatial template can be viewed as a model of how objects are approximately positioned and oriented relative to other objects. These relative locations and orientations are specified through constraints on object attributes.

**Definition 1** (Spatial object) *A spatial object is an object characterized by a set of attributes, two of which represent measures of object location and orientation in space.*

**Definition 2** (Spatial template) *A spatial template T (also referred as a template) is a structure $\langle A, O, C \rangle$, where*

- $A = \{A_1, A_2, \ldots, A_k\}$ *is a set of attributes, including location and orientation in space, each attribute $A_i$ has a range of values $R(A_i)$ corresponding to the set of values it can take;*

- $O = \{O_1, O_2, \ldots, O_l\}$ *is a set of spatial objects characterized by $A$;*

- $C = \{C_1, C_2, \ldots, C_m\}$ *is a set of constraints (relations), including fuzzy constraints, defined over subsets of $O$.*

**Definition 3** (Instance of a spatial object) *An instance of a spatial object $O_i$ is a specific set of values taken by the attributes of $O_i$ and denoted $o_i = \{a_1, a_2, \ldots, a_k\}$,*

$a_j \in R(A_j)$, $j = 1, 2, \ldots, k$, *where k is the number of attributes. The set of all possible spatial object instances is the Cartesian product of all attribute ranges and is*

*denoted $I(O) = \prod\limits_{j=1}^{k} R(A_j)$.*

From now on, we will refer to spatial objects simply as objects. The first two definitions formally state that a spatial template is a collection of objects located and oriented in space, with other characteristics, and spatially arranged following some constraints. Though we will be dealing with templates whose location attribute range is the 2-dimensional space, it should be clear that the concept also applies for any n-dimensional space (n≥2). Particular applications may not be concerned with the spatial orientation of objects, in such cases, the orientation attribute can be left out. The other undefined attributes will usually be associated with characteristics that are application dependent and, for example, could refer to types or categories of objects[1].

The set of constraints is used to specify relations between template objects as well as object characteristics. As we will see in the application described hereafter, we are mostly concerned with constraints on the relative locations and orientations of objects although the constraints might also apply to other attributes of objects.

### 2.2   FUZZY CONSTRAINTS IN TEMPLATES

A template can be used to represent many possible arrangements of its object constituents. These arrangements are specified through the set of template constraints. In the usual sense, a constraint is a condition that must be satisfied by the instantiated objects to which it is applied. However, to allow more flexibility, it is useful to consider constraints that can be satisfied partially. To do so, we allow the set of template constraints to include fuzzy constraints as well. For our purpose, we will use the following definition for a fuzzy constraint — we assume the reader is familiar with some basic concepts such as fuzzy set and fuzzy relation (for an introduction to fuzzy set theory, we refer to Kaufmann (1975)).

**Definition 4** (Fuzzy constraint) *A fuzzy constraint $C_i$ of a given template $\langle A, O, C \rangle$ is a fuzzy relation of arity r ($r \leq Cardinality(O)$) over $I(O)^r$ (r-fold Cartesian product of $I(O)$).*

---

[1] The notions of attributes and objects are very similar in this context to the notions of features and patterns in pattern recognition in the sense that attribute ranges define a space where object instances are points in the same way features define a feature space in which patterns are points (or feature vectors). However, the problem as stated here is not the recognition of object instances but of arrangements of object instances. Thus, patterns are not object instances but object arrangements in space.

This definition states that a fuzzy constraint $C_i$ implies $r$ template objects related in some way, and associates a membership function in $[0,1]$ to this relation. Hence, a fuzzy constraint is a fuzzy subset of $I(O)^r$ whose elements consist of $r$-tuples of object instances. The fuzzy constraint is partially satisfied by each of these tuples according to their membership value with $1$ being the maximum satisfaction achievable (fully satisfied), and $0$ the minimum satisfaction (not satisfied).

Let us consider the following example of a spatial template in 2-dimensional space (XY-plane) which uses both fuzzy and non-fuzzy constraints.

**Example 1**

$T = \langle A, O, C \rangle$ where

$A = \{Loc, Orien, Color, Size\}$ is the set of attributes with ranges

$R(Loc) = \mathbf{R}^2$ (coordinates in the XY-plane)

$R(Orien) = [0, 360]$ degrees (orientation)

$R(Color) = \{blue, red, green, yellow\}$

$R(Size) = \{small, medium, big\}$

$O = \{O_1, O_2, O_3, O_4\}$ is the set of objects having attributes $Loc, Orien, Color, Size$.

$I(O) = \mathbf{R}^2 \times [0, 360] \times \{blue, red, green, yellow\} \times$

$\{small, medium, big\}$

$C = \{C_1, C_2, C_3, C_4, C_5\}$ is the set of constraints with

$C_1 = \{(o_1):(o_1) \in I(O), o_1 = (\_,\_, red, big)\}$ [2]

$C_2 = \{(o_2):(o_2) \in I(O), o_2 = (\_,\_, blue, small)\}$

$C_3 = \{(o_1, o_3):(o_1, o_3) \in I(O)^2, o_1 = (loc_1, \_, \_, \_),$

$\quad o_3 = (loc_3, \_, \_, \_), distance(loc_1, loc_3) \le 6\}$

$C_4 = \{(o_2, o_3, o_4) / \chi_{C_4}(o_2, o_3, o_4):(o_2, o_3, o_4) \in I(O)^3,$

$\quad o_2 = (loc_2, \_, \_, \_), o_3 = (loc_3, \_, \_, \_),$

$\quad o_4 = (loc_4, \_, \_, \_), \chi_{C_4}(o_2, o_3, o_4) \in [0,1]\}$

where $\chi_{C_4}(o_2, o_3, o_4)$ is a specified measure of the alignment of points $loc_2$, $loc_3$ and $loc_4$.

$C_5 = \{(o_2, o_4) / \chi_{C_5}(o_2, \bullet_4):(o_2, o_4) \in I(O)^2,$

$\quad o_2 = (\_, orien_2, \_, \_), o_4 = (\_, orien_4, \_, \_),$

$\quad \chi_{C_5}(o_2, o_4) \in [0,1]\}$

where $\chi_{C_5}(o_2, o_4)$ is a specified measure of the closeness of $orien_2$ and $orien_4$.

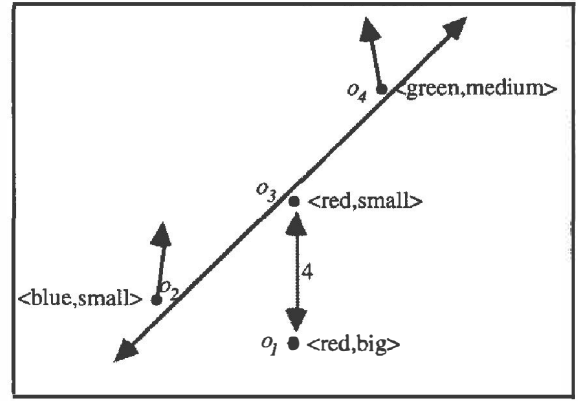

Figure 1: Example of a Spatial Template Instance

This example includes three non-fuzzy constraints: $C_1$ states that object $O_1$ must be red and big (it does not constrain the other attributes), $C_2$ states that object $O_2$ must be blue and small, and $C_3$ states that the distance between objects $O_1$ and $O_3$ must not exceed 6. The other constraints are fuzzy constraints: $C_4$ states that objects $O_2$, $O_3$ and $O_4$ must be approximately aligned, and $C_5$ states that $O_2$ and $O_4$ should have similar orientations. Both fuzzy constraints can be partially satisfied by an infinite number of object instances with satisfaction values defined by $\chi_{C_4}$ and $\chi_{C_5}$ respectively.

## 2.3  SPATIAL TEMPLATE INSTANCES

As illustrated in the previous example, a template is really a class where the set of constraints can possibly be satisfied by an infinite number of different object instances. This leads to the definition of a template instance.

**Definition 5** (Template instance) *Let* $T = \langle A, O, C \rangle$ *be a spatial template where* $A = \{A_1, A_2, ..., A_k\}$, $O = \{O_1, O_2, ..., O_l\}$ *and* $C = \{C_1, C_2, ..., C_m\}$, *an instance* $t$ *of* $T$ *is an element of* $I(O)^l$ *(i.e., a $l$-tuple of instantiated objets) such that for every* $i \le m$,

- *if* $C_i$ *is a non-fuzzy constraint then it is satisfied by* $t$;

- *if* $C_i$ *is a fuzzy constraint then it is at least partially satisfied by* $t$, *i.e.,* $\chi_{C_i}(t) > 0$.

Informally, a template instance is a particular set of object instances which satisfies all the template constraints[3]. An interesting characteristic of template instances is that they can be rotated and translated to give other instances of the same template; thus templates are invariant under rotation and translation. Figure 1 shows an instance of the template in Example 1.

A closely related notion to template instance is that of a spatial situation (also referred hereafter as a situation).

**Definition 6** (Spatial situation) *A spatial situation is a set of objects with specific values for the location attributes. Other attributes may or may not have defined values.*

The main difference between a template instance and a spatial situation is that the spatial situation does not necessarily include object instances since only the location attribute must have a specific value. Besides, a spatial situation will usually have many more elements than a template instance that will be matched upon it, as will be described in section 4. Spatial situations serve as inputs for the recognition of template instances. Obviously, the object attributes of a spatial situation must include the attributes of any template in which we wish to recognize instances.

## 2.4 MAIN ISSUE RELATED TO TEMPLATE REPRESENTATION

We have reduced the task of representing spatial templates to the task of specifying fuzzy constraints on template objects. In real applications, and as illustrated by Example 1, constraints are not all independent, that is, an object can be bound by many constraints. In addition, it will be useful and even necessary, as we will see, to define constraints hierarchically, so that constraints can be defined in terms of other constraints. As a consequence, the problem of representing spatial templates becomes very tough for complex templates containing many objects. The main issue investigated in this work is to develop an expressive and meaningful model to represent or specify dependent fuzzy constraints where the fuzzy variables are location and orientation.

In the following section, we will be dealing mostly with location and orientation attributes — fuzzy constraints will be defined for these attributes as they are the main concern of this paper. Other attributes need only be considered as object types which are specified by non-fuzzy constraints — they will be omitted hereafter.

## 3 REPRESENTATION MODEL

Spatial templates are highly geometric in nature. Therefore, a good way to allow the specification of complex templates is by defining them through a set of hierarchically structured geometric elements or building blocks. To simplify the specification of fuzzy constraints, we will use fuzzy geometric relations (FGRs). FGRs will provide a means of expressing the fuzzy constraints on location and orientation as nested relations. The first subsection describes the FGRs that have been used in the particular domain in which the results of this work have been applied (see section 5 for more details about the application). Other FGRs can be added as needed in a particular domain. The second subsection describes the way fuzzy constraints are specified.

## 3.1 FUZZY GEOMETRIC RELATIONS

A fuzzy geometric relation is an arrangement of oriented reference points in space that follows a specific oriented geometric pattern or figure. Although it may seem strange to talk about point orientation, let it be clear that, in this context, a reference point is used to refer to an oriented structured object which can be another oriented geometric pattern itself. The fuzziness of a FGR is specified through a membership function that indicates the "closeness" of the arrangement of a given ordered set of reference points with the geometric pattern involved.

A proximity relation is defined as a fuzzy relation which is reflexive, symmetric, but not necessarily transitive (Lee, 1972). Lee used the proximity relation as a quantitative measure of the proximity of two n-sided polygons. In this work, proximity relations are used to assess the degree of likeness of an arrangement formed by an ordered tuple of points with a subset of standard geometric figures, namely, isosceles triangle, equilateral triangle, rectangle triangle and rectangle. In addition, proximity relations are used to represent other fuzzy spatial relations between points, namely, fuzzy ring sector, fuzzy trapezoidal section and fuzzy alignment.

Two aspects are taken into consideration in order to allow variability in fuzzy geometric relations. The first aspect, which only applies to the geometric figures, relates to the general shape similarity. The second focuses on the variation or elasticity of the dimensions and orientations considered (base, height, side, distance and relative orientation). The first aspect is modeled by defining a proximity relation based on angles. The definitions of the proximity relations used in this work are very similar to those used by Lee (1978). Table 1 captures this idea[4] — the proximity measures of the four geometric figures are invariant with respect to expansion and contraction[5]. The second aspect (dimension and orientation) is considered by the use of fuzzy sets. Table 2 gives the fuzzy sets for each of the seven FGRs considered. As an example, a fuzzy isosceles triangle has five fuzzy sets: base, height, relative orientations of vertices $A$, $B$ and $C$ (the orientation of the vector normal to the vector $AB$ is taken as the reference orientation).

---

[4]We call proximity measure, the fuzzy value associated to a proximity relation. $\angle X$ denotes the angle at vertex $X$ and $\|XY\|$ the length of the vector going from $X$ to $Y$.

[5]Dubois and Jaulent (1985) have also defined very interesting proximity measures for various primitive shapes. Their set of shapes is more exhaustive than ours.

Table 1: Proximity Measure for the General Shape of Geometric Figures

| Geometric Figures | | Proximity Measures |
|---|---|---|
| Isosceles triangle | 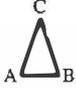 | $1 - \dfrac{|\angle A - \angle B|}{\pi}$ |
| Equilateral triangle | 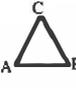 | $1 - \dfrac{\left|\angle A - \pi/3\right| + \left|\angle B - \pi/3\right| + \left|\angle C - \pi/3\right|}{4\pi/3}$ |
| Rectangle triangle | 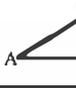 | $1 - \dfrac{\left|\angle B - \pi/2\right|}{\pi/2}$ |
| Rectangle | 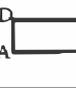 | $1 - \dfrac{\left|\angle A - \pi/2\right| + \left|\angle B - \pi/2\right| + \left|\angle C - \pi/2\right| + \left|\angle D - \pi/2\right|}{2\pi}$ |

Table 2: Fuzzy Geometric Relations and their Fuzzy Sets

| Fuzzy Geometric Relations | Fuzzy Sets |
|---|---|
| Fuzzy isosceles triangle $<A,B,C>$ | Base, height, relative orientations of $A$, $B$ and $C$. |
| Fuzzy equilateral triangle $<A,B,C>$ | Side, relative orientations of $A$, $B$ and $C$. |
| Fuzzy rectangle triangle $<A,B,C>$ | Base, height, relative orientations of $A$, $B$ and $C$. |
| Fuzzy rectangle $<A,B,C,D>$ | Base, height, relative orientations of $A$, $B$, $C$ and $D$. |
| Fuzzy ring sector $<A,B>$ | Distance $\|AB\|$, relative orientation of vector $AB$, relative orientation of $B$. |
| Fuzzy trapezoidal section $<A,B>$ | Projected distance $\|AB\|$, relative orientation of vector $AB$, relative orientation of $B$. |
| Fuzzy alignment $<A_1,A_2,...,A_n>$ | Projected distances $\|A_1A_2\|$, $\|A_2A_3\|$, ..., $\|A_{n-1}A_n\|$, relative orientation of alignment, relative orientations of $A_1, A_2, ..., A_n$. |

The proximity measure of a given tuple of oriented points with a given FGR is simply defined as the minimum of the proximity measure for the general shape (this only applies to the geometric figures) and all the membership degrees of observed values in the fuzzy sets used to define the FGR. As an example, the proximity measure of a given triple of points $<P_1,P_2,P_3>$ with a fuzzy isosceles triangle (defined by the fuzzy sets *base, height, orien_A, orien_B and orien_C*) is the minimum of the six following measures:

$$\omega, \mu_{base}(\|P_1P_2\|), \mu_{height}(\|P_3, proj\_P_3\|),$$
$$\mu_{orien\_A}(rel\_orien\_P_1), \mu_{orien\_B}(rel\_orien\_P_2),$$
$$\mu_{orien\_C}(rel\_orien\_P_3)$$

where $\omega$ is the general shape proximity measure of the triple $<P_1,P_2,P_3>$ with the isosceles triangle as defined in Table 1, $\mu_{base}(\|P_1P_2\|)$ is the membership degree of the distance between points $P_1$ and $P_2$ in the fuzzy set *base*, $\mu_{height}(\|P_3, proj\_P_3\|)$ is the membership degree of the computed height from $<P_1,P_2,P_3>$ in the fuzzy set *height*, and $\mu_{orien\_A}(rel\_orien\_P_1)$, $\mu_{orien\_B}(rel\_orien\_P_2)$ and $\mu_{orien\_C}(rel\_orien\_P_3)$ are the membership degrees of the relative orientations of $P_1$, $P_2$ and $P_3$ in the fuzzy sets $orien\_A$, $orien\_B$ and $orien\_C$, respectively. The proximity measures for equilateral triangle, rectangle triangle and rectangle are defined similarly.

Figures 2 and 3 are illustrative examples of two shapeless FGRs, that is, fuzzy trapezoidal section and fuzzy ring sector, respectively. Lighter areas correspond to higher proximity measures than darker areas. A fuzzy ring sector is used to constrain the distance between the two objects $A$ and $B$, as well as the angle they form with respect to object $A$ orientation. Therefore, the following approximate concepts can be captured: *to the left, to the right, in front* and *behind*. A fuzzy trapezoidal section is similar to a fuzzy ring sector but the observed distance is projected on the segment that maximizes the membership degree in the fuzzy set used to define the relative orientation of vector $AB$.

The fuzzy alignment relation is specified by an ordered list of objects along with their respective orientations, respective adjacent distances, and an additional relative orientation for the fitted straight line through the objects.

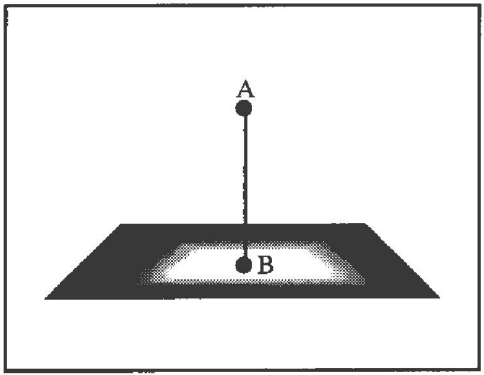

Figure 2: Fuzzy Trapezoidal Section

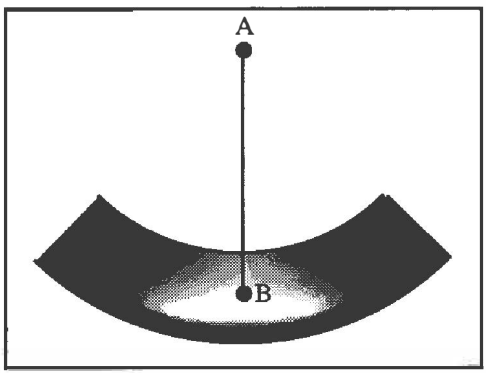

Figure 3: Fuzzy Ring Sector

The fitting criteria used here is the least square fit.

## 3.2 SPECIFICATION OF FUZZY CONSTRAINTS

A fuzzy constraint is specified as a FGR that applies to reference points in two-dimensional space. A reference point represents either the location and orientation of an object, or the reference coordinates and orientation associated with another FGR. Constraints can thus be viewed as nested FGRs. Hence, the set of template constraints defines a hierarchy of FGRs in which the same FGR (or object) can be used by several constraints. This hierarchical structure corresponds formally to a directed acyclic graph (DAG) as illustrated by Figure 5.

The reference point orientation is only determined by the kind of FGR, for instance, the reference orientation of a fuzzy triangle is the vector normal to the triangle base while the reference point orientation of a fuzzy ring sector is the orientation of vertex $A$. On the other hand, there are several choices for the reference point coordinates of a given FGR. For instance, it could be the center of mass of the locations of all objects directly or indirectly used in the arguments of the FGR, or the center of mass of the reference point coordinates (or object locations) of all arguments of the FGR. Other possibilities are the location of the upper, lower, leftmost or rightmost object directly or indirectly implied by the FGR with regard to the FGR orientation. When specifying a fuzzy constraint, the way to select the reference coordinates is specified as well. Lower level FGRs must be instantiated before higher level FGRs can be instantiated. Once a FGR has been instantiated, the reference points needed by its parent FGRs can be computed.

## 4    RECOGNITION OF TEMPLATE INSTANCES

### 4.1   RECOGNITION ALGORITHM

A spatial template can be matched against a spatial situation to look for particular instances in the situation. The recognition algorithm is given a modeled template, a spatial situation and a threshold value. Of course, the matching process is possible only if the attributes of template objects are also attributes of the situation objects. The general matching process consists in establishing a bijection (pairing) between template object instances and a subset of objects in the situation such that constraints are satisfied (note that paired objects must be of the same type). More precisely, it consists in instantiating FGRs in a bottom-up manner over the hierarchical structure of FGRs. First, a FGR that is defined on objects only is picked up, and then these objects are paired with a combination of situation objects. If the proximity measure associated with this combination is over the threshold value then the procedure tries to instantiate the parent FGRs, otherwise another combination is tried (this is known as a "cut in the search path"). This procedure stops when all the root FGRs are instantiated, that is, all template constraints are satisfied with corresponding proximity measures over the threshold value.

Any undefined attribute of situation objects might be assigned a value by the recognition process so that the situation object type corresponds to the type of the template object instance. In addition, in the implementation, we allow object instances to have undefined attributes (except location). The proximity measure of a constraint is set to the minimum over the proximity measures of all FGRs implied in the constraint. Once an instance is found, the resulting template proximity measure is set to the minimum over the proximity measures of all template constraints. The output of the recognition algorithm consists in the template instances found along with the corresponding proximity measures and situation object attributes that have been assigned values. Details of the recognition algorithm can be found in the contract technical report (APG Inc., 1994).

The representation model and recognition process have been implemented in Quintus Prolog on a SPARC Station IPC. We benefited from some aspects of logic programming and, more particularly, the Prolog language. First, the language syntax is well suited for our representation model — predicates are used to represent fuzzy relations. Second, the backtracking mechanism embedded in the Prolog inference engine relieved us from designing a non-deterministic algorithm. This allows recognition of multiple instances and backtracking to previous choices in the search space whenever a cut or failure occurs.

### 4.2 COMPLEXITY ANALYSIS

The structure of the hierarchy of FGRs is a directed acyclic graph (DAG). Contrary to many graph search algorithms, the goal of the recognition process is not to find a particular path in the graph but to estimate a function (proximity measure) at each node.

Let $m$ be the number of template objects, $k$ the number of FGRs in the template and $n$ the number of situation objects. For a specific template ($m$ and $k$ fixed), the complexity of the recognition process depends on the number of arrangements of $m$ object in $n$, therefore, it is polynomial time in $n$, that is,

$$km! \binom{n}{m} = k \frac{n!}{(n-m)!} \cong O(n^m).$$

However, in the general case, an instance of the problem has size dependent on $n$, $m$ and $k$, therefore the problem is clearly NP-hard since it is $O(kn^m)$ where $m$ and $k$ are not fixed.

## 5  APPLICATION

This work was initially motivated by a military problem related to the identification, from imprecise observations, of military formations deployed on the field[6]. The uncertainty associated with observed data location and orientation is not provided as input. However, doctrinal deployment templates, which represent the ways given formations (composed of many different units) typically deploy when performing specific activities, can take uncertainty into account in their representation. In the current application, a template object represents a unit. A unit is located in the field, and follows (if it is moving) or faces in a direction — this defines the object location and orientation. There are several types of units although they are not specified hereafter. A spatial template is a representation of a specific doctrinal deployment template for a generic formation (for instance, a model of how any tank regiment typically deploys when advancing). A spatial situation merely consists in the set of observed units (also called tactical situation), or a subset of them. The proximity measure associated with a recognized template instance is interpreted as a measure of "goodness" of the found match or simply as a degree of confidence.

It is essential to take into account the various possible distortions to deployment templates such as rotation, expansion, compression and local compression. Furthermore, a hierarchical representation model is required as the problem is inevitably hierarchical, for instance, a division is composed of regiments which are composed of battalions.

### 5.1  A COMPLEX EXAMPLE

Doctrinal deployment templates in military books are not rigorously specified. The concept of echelons is used to draw horizontal boundaries between the different parts of template — vertical boundaries are also used to delimit template extent. Relative locations of units are specified by these boundaries. For instance, a range is given for the distance between a unit (in the first echelon) and the front line. In our representation model, the concept of boundaries is left out. Instead, units are grouped and their relative locations specified by use of geometric properties.

---

[6]An important and difficult part of an intelligence analyst's task is to analyze spatial relationships between sighted units to aggregate them into higher level formations and then recognize their main activity (DMR Group Inc., 1992). Besides, the analyst has to hypothesize the presence of units not yet observed. Matching of doctrinal deployment template against a given tactical situation is the basic method used to achieve this. This particular application is aimed at providing an automated tool to help the analyst in recognizing deployment template instances in a given situation.

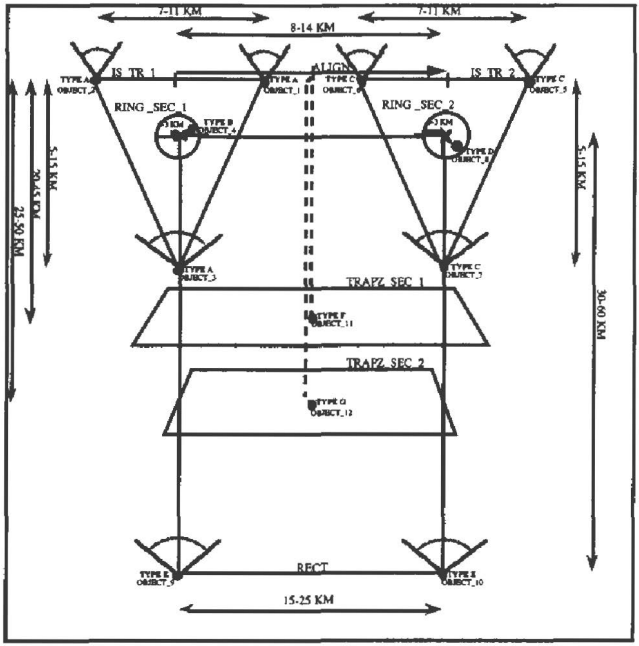

Figure 4: Example of a Modeled Deployment Template

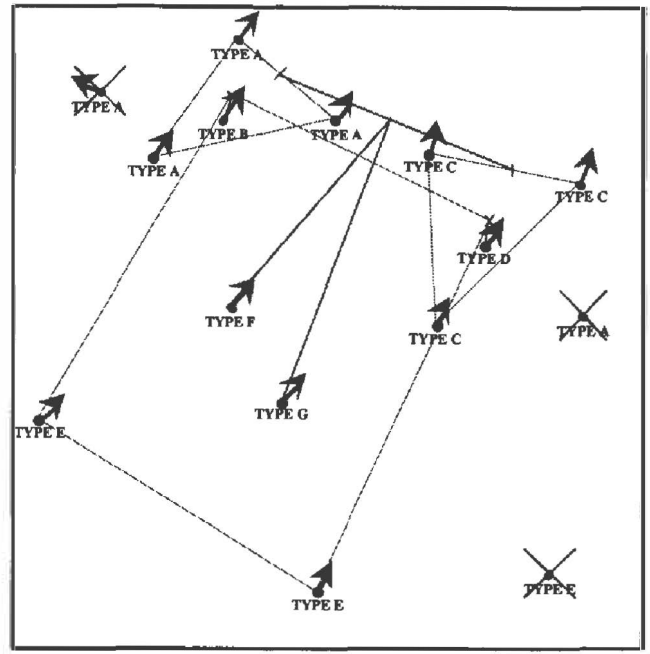

Figure 6: A Recognized Template Instance in a Given Tactical Situation

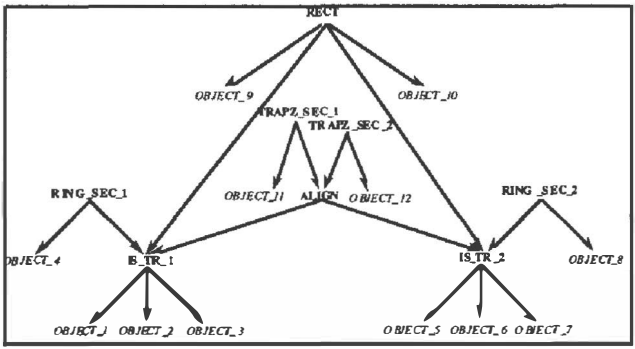

Figure 5: Hierarchy of FGRs for the Example Template.

Figure 4 illustrates a doctrinal deployment template represented in accordance with our model. It represents a motorized rifle division in attack according to the *Fantasian Ground Forces* (organizational guide). The template is composed of twelve objects (units) and eight fuzzy geometric relations are built upon these objects: two isosceles triangles, two ring sectors, two trapezoidal sections, one alignment and one rectangle. The hierarchy of FGRs (DAG) is shown in Figure 5 (we refer to Figure 4 for the names of the constraints). The FGR *IS_TR_1* is used by three other FGRs, namely, *RING_SEC_1*, *ALIGN* and *RECT* — selected reference point locations are, respectively, center of mass of the three objects, center of mass of the two objects that form the triangle base, and center of mass of the three objects. Both trapezoidal sections (*TRAPZ_SEC_1* and *TRAPZ_SEC_2*) are defined in terms of the center of mass of the two reference point locations used for the alignment *ALIGN*.

We provided our system with the definition of the template modeled in Figure 4 using appropriate fuzzy sets. In addition, the tactical situation shown in the background of Figure 6 (not exactly to scale) was used as input. The situation contains fifteen units with seven different types. Two template instances were found in the situation in approximately seven seconds. The first instance, with proximity measure 0.83, is shown in Figure 6. The second instance is very similar to the first one — only the lower right object of type *E* in the situation is substituted for the lower right vertex of the fuzzy rectangle (the proximity measure is the same as with the first instance since the FGR with minimum proximity measure is *IS_TR_1* in both cases).

# 6  FURTHER WORK

Although the complexity of the general problem will remain NP-hard, there are ways to speed up the recognition algorithm by searching more wisely the space of potential template instances. At first glimpse, it seems that we could benefit more from the spatial aspect of the problem. For instance, we could initially create a table that gives, for each object in the situation, the list of its *K* nearest neighbors (objects) along with their distance to the given object. Such calculation can be done efficiently (polynomial time) using graph theoretic methods and has to be carried out only once for a given situation. Once this table is available, it can be used, for example, to filter out potential object combinations whose span exceeds the span of the FGR being matched.

Since spatial templates are specified by fuzzy constraints, the problem of recognizing template instances can be seen as a partial constraint satisfaction problem (PCSP) where *partial* can be interpreted as *fuzzy*. We might benefit from PCSP techniques. This is something we want to look at in future work to improve the search strategy.

We want to carry out an interesting extension to our method that will allow us to cope with the problem of missing objects in the situation. In the application discussed in section 5, it is common for the situation to include "holes", that is, some units (or objects) are not observed. In such a case, with the actual algorithm, no template instance can be recognized. We want to study techniques to recognize incomplete instances, i.e., instances in which a few objects are missing. In addition, distribution of possibility, or the best location with respect to proximity measure, for the missing objects should be provided. One possibility to solve in part this problem is to allow at most one missing object for a geometric figure to be recognized.

# 7  CONCLUSION

We have introduced the concept of spatial templates to represent generic arrangements or patterns of interrelated objects in space. We have reduced the problem of representing templates to the problem of specifying fuzzy constraints on objects. We have defined a set of fuzzy geometric relations, which can be hierarchically structured, to simplify the specification of fuzzy constraints on object locations and orientations. We have implemented an algorithm that recognizes template instances from observed data, and associates proximity measures to these instances. The proximity measure represents the degree of similarity of an instance with its template. The results of this work have been applied to a complex and interesting problem related to the representation of military deployment template. The representation model has been shown to be very expressive and meaningful. Moreover, the representation of hierarchically structured groupings turned out to be essential in this domain. We believe our work may have influence in several areas such as (fuzzy) pattern recognition and computer vision.

## Acknowledgments

The authors are grateful to members of the DREV Combat Automation Group and the DMR team for sharing their ideas during project initiation. Special thanks are addressed to Steve Woods who provided fruitful comments and worked on the same problem using a different approach. We also greatly appreciated the military expertise of Gary Handson for the application of our results. We also thank Claude Proulx for proofreading the text.